# DESIGN CONSIDERATIONS FOR HYPOTHESIS REJECTION MODULES IN SPOKEN LANGUAGE UNDERSTANDING SYSTEMS

*Aman Alok, Rahul Gupta, Shankar Ananthakrishnan*

Amazon.com, Cambridge, MA-USA

## ABSTRACT

Spoken Language Understanding (SLU) systems typically consist of a set of machine learning models that operate in conjunction to produce an SLU hypothesis. The generated hypothesis is then sent to downstream components for further action. However, it is desirable to discard an incorrect hypothesis before sending it downstream. In this work, we present two designs for SLU hypothesis rejection modules: (i) scheme R1 that performs rejection on domain specific SLU hypothesis and, (ii) scheme R2 that performs rejection on hypothesis generated from the overall SLU system. Hypothesis rejection modules in both schemes reject/accept a hypothesis based on features drawn from the utterance directed to the SLU system, the associated SLU hypothesis and SLU confidence score. Our experiments suggest that both the schemes yield similar results (scheme R1: 2.5% FRR @ 4.5% FAR, scheme R2: 2.5% FRR @ 4.6% FAR), with the best performing systems using all the available features. We argue that while either of the rejection schemes can be chosen over the other, they carry some inherent differences which need to be considered while making this choice. Additionally, we incorporate ASR features in the rejection module (obtaining an 1.9% FRR @ 3.8% FAR) and analyze the improvements.

## 1. INTRODUCTION

Spoken Language Understanding (SLU) systems have become ubiquitous with the introduction of personal assistants such as Amazon Alexa, Google Home, Microsoft Cortana and Apple Siri [1, 2]. A typical design for an SLU system consists of an ASR component, pipelined with a language understanding component. The SLU system yields the hypotheses in a predefined schema – for instance, an utterance decoded as "what is the time in Boston" by ASR, could be hypothesized as Time—GetTimeIntent—Boston:City by the SLU system. The predicted hypothesis is sent downstream and a response is returned to the user. However, acting on the top hypothesis generated by the SLU system may not always be desirable. For instance, in case of the utterance above, if the SLU system returns PlayMusic as the top hypothesis, it is more sensible to reject this particular hypotheses instead of passing it to the downstream components. We argue that in such cases a *recovery question* or even *no response* is better than the acting on the incorrect hypothesis. In this work, we present considerations for the design of hypothesis rejection modules. Specifically, we present two design schemes for hypothesis rejection - an early hypothesis rejection scheme that acts on domain specific hypotheses list and, the second scheme that performs hypothesis rejection at the overall SLU system level. We describe the training and evaluation methodologies for the two schemes and list out their characteristics. We recommend that these properties be considered while choosing one of the two schemes.

SLU systems [3, 4] are integral components of conversational agents. Several design schemes have been proposed for SLU systems with applications of standard machine learning algorithms such as active learning [5], semi-supervised learning [6] and discriminative/generative modeling [7, 8]. Tuning rejection in a machine learning system is a basic operation and it is common to find a constant rejection threshold to apply on the system confidence that yields a desired False Accept Rate (FAR) or False Reject Rate (FRR) [9]. This standard method can also be applied to SLU systems, however we argue that rejection in SLU systems can be performed based on richer signals instead of just thresholding the SLU confidence score. Previous works have sourced multiple signals apart from the utterance text fed to the SLU systems to enhance the SLU model performance. Examples include the use of prosodic signals [10], features extracted from ASR components [11, 12] and other multi-modal features [13]. In this work, we explore if a larger set of features used for generating SLU confidence scores can also be used to inform hypothesis rejection (in our work we use features representations obtained on the utterance and SLU hypothesis). To the best of our knowledge, this is the first work exploring multiple schemes for hypothesis rejection in SLU systems. In the proposed schemes, machine learning models are trained on a dataset (separate from the one used in SLU model training) to learn patterns where a generated SLU hypothesis can be potentially incorrect.

We explore two schemes for hypothesis rejection: (i) an early rejection within domain specific components in a SLU system, followed by pooling of the accepted hypotheses and, (ii) a late hypothesis rejection post pooling of all domain specific hypotheses, at the overall SLU system level. Given a modeling architecture, we discuss the training and evaluation methodologies in both the cases. For each scheme, we explore the value of adding more signals to inform hypothesis rejection beyond just the SLU system confidence score (as would be done in a conventional machine learning setup). In this work, we add the utterance under consideration and the SLU system hypothesis. The early rejection scheme obtains a False Reject Rate (FAR) of 2.5% @ 4.5% False Accept Rate (FAR), while the late rejection schemes obtains a FRR of 2.5% @ 4.6% FAR. We discuss the advantages and disadvantages of both the schemes and present mechanisms to allow control of FAR/FRR operating points. Finally, we also conduct experimentation on a hypothesis rejection module operating on an SLU system operating on ASR generated text (obtaining an FRR of 1.9% @ 3.8% FAR). We further perform an analysis on the errors recovered by the rejection module after addition of ASR features to the rejection modules.

## 2. BACKGROUND ON SLU SYSTEM

Figure 1 depicts the structure of the SLU system used in our analysis. The architecture is borrowed from Su et al. [14], and consists of four set of models for each domain supported by the SLU system: (i) a Domain Classifier (DC), (ii) an Intent Classifier (IC), (iii) a Named Entity Recognizer (NER) and, (iv) a reranker. The DC returns the confidence score for an utterance to be belonging to that domain (for instance in the Figure 1, the utterance "play moana" is assigned a score of 0.6 by music domain classifier, and 0.7 by books domain classifier). The IC returns confidence scores for a set of in-tents within a given domain (for instance in the Figure 1, the example utterance is assigned scores for ReadBookIntent, FindBookIntent and BuyBookIntent within the Books domain). The NER model tags entities within the utterance (for instance in Figure 1 moana is tagged as BookName and PageNumber with confidence of 0.4 and 0.2, respectively). A set of hypotheses is created for each domain (as shown in Figure 2) by performing a Cartesian product over the outputs from DC, IC and NER models. For each hypothesis, an associated score is returned by the reranker based on the DC, IC and NER confidence scores. The reranker returns the SLU confidence score and is depicted as f(DC output, IC output, NER output) in the Figure 1. A detailed description of these models, the training procedure and the hypothesis generation process can be found in [14].

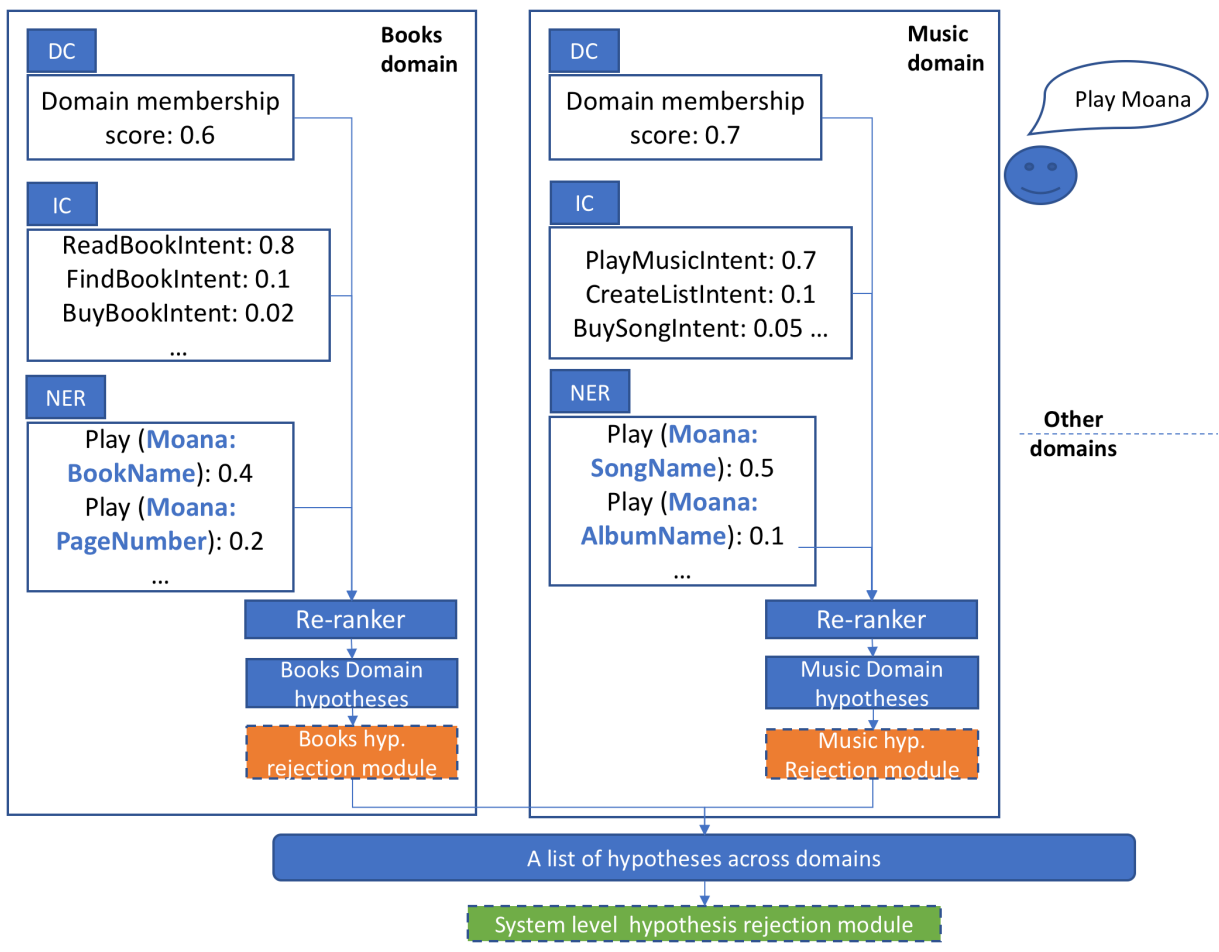

**Fig. 1**. Setup of the SLU system used in our experiments (borrowed from [14]). The domain specific rejection modules are depicted in orange and they reject hypothesis per domain. Alternately, an overall system level rejection module is shown in green.

## 3. HYPOTHESIS REJECTION MODULES DESIGN

Given the architecture of the SLU system as described above, we consider two design options for the hypothesis rejection modules: (i) R1: A domain specific hypothesis rejection scheme and, (ii) R2: a SLU system level hypothesis rejection scheme. We describe the two design schemes in more detail below.

### 3.1. R1: Domain specific rejection modules

A domain specific hypothesis rejection design is depicted by orange blocks in the Figure 1. This scheme requires a rejection module for each domain, operating on hypotheses for the corresponding domain. Given a set of domain specific hypotheses, the domain's rejection module returns a subset as candidates from that domain. Thereafter, accepted set of hypotheses from each domain are pooled to create a cross-domain list of hypotheses. The accepted SLU hypotheses are then ranked based on the SLU scores after pooling.

### 3.2. R2: SLU system level rejection module

In this scheme, we first pool the SLU hypotheses and perform rejection on the list of hypotheses merged from all the domains. Given the pooled set of cross-domain hypotheses, the rejection module returns a subset as candidates to downstream components. This module is represented by an green block in the Figure 1.

Note that while there are as many rejection modules as the domain count for the scheme R1, we have only one rejection module for the scheme R2. This comparison is akin to a comparison between multi-class and one-vs-all classification setup. Whereas the R2 scheme has to reject hypotheses pooled across all the domains, a given R1 rejection module is only concerned with rejecting incorrect hypotheses from the corresponding domain. We also note that while the rejection modules trim the generated set of SLU hypotheses to yield a shorter hypotheses list, SLU systems typically operate on the top hypothesis (the hypothesis with the highest SLU score in the accepted list). Going forward with this assumption, we train and evaluate our rejection modules on top hypothesis from each domain in case of R1 and the overall top SLU hypothesis in case of R2.

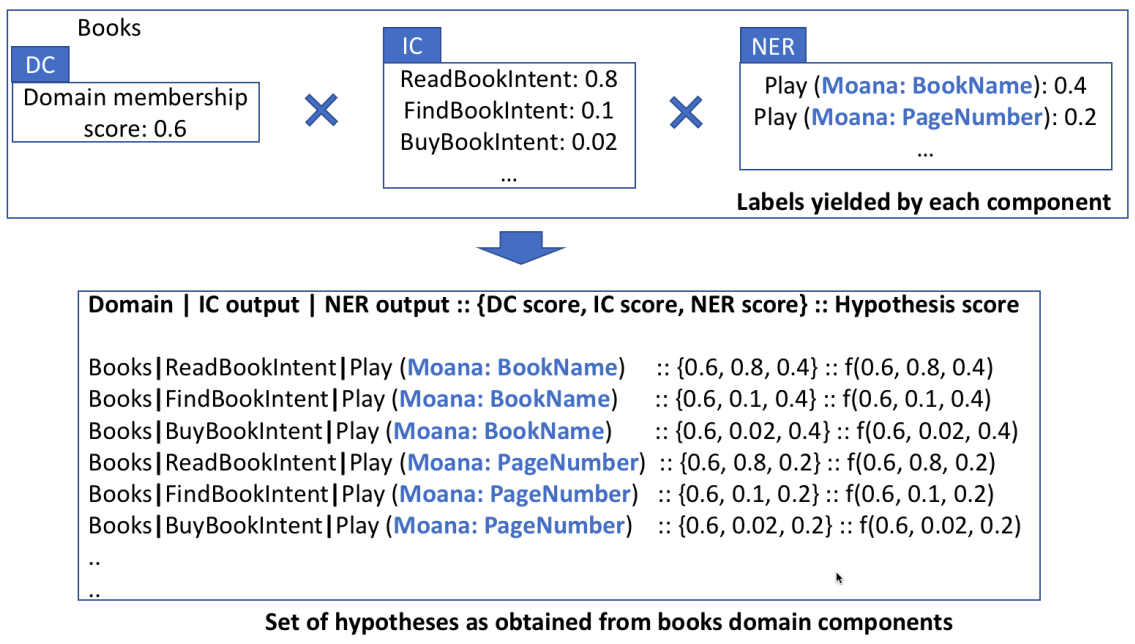

**Fig. 2**. Hypotheses generation using models for the Books domain.

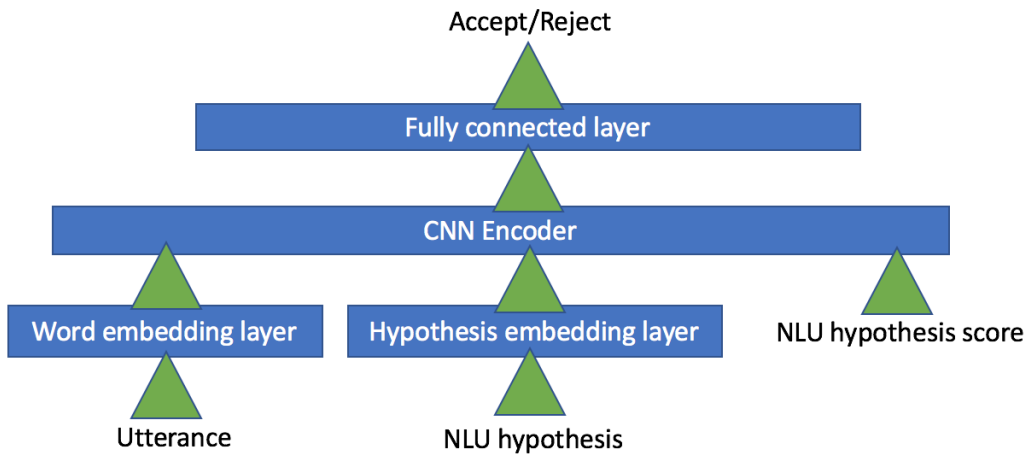

**Fig. 3**. Modeling architecture used for our rejection modules. The figure shows all three features used, but we also experiment with a subset of these features.

## 4. MODELING SETUP

The high level architectural design of the rejection modules is shown in Figure 3. In our experiments, the input features to the models is a subset of following: the utterance text (e.g. "play moana"), the corresponding hypothesis being considered (e.g. Books—ReadBookIntent—BookName) and the corresponding SLU score (as returned by the reranker: e.g. f(0.6, 0.8, 0.4)). The model first projects the input text and the SLU hypothesis into embeddings (individual embeddings are learnt for each word, as well as each individual component of the SLU hypothesis: domain, intent and slots). The word and SLU hypothesis embeddings are combined with the SLU interpretation score and passed through a Convolutional Neural Network (CNN) encoder [15], returning a joint representation for the utterance, SLU hypothesis and the SLU score. The word and hypothesis embeddings are learnt by backpropagating the errors through these parameters while training the hypothesis rejection models. Finally, a fully connected layer returns decision as to accept/reject a particular SLU hypothesis. We use a cross-entropy objective to train the model parameters and model configurations (dimensionality of word and hypothesis embeddings, number of CNN filters, number of fully connected layers etc.) are fine tuned on the development set.

### 4.1. Training methodology

Although the model architecture is similar for the two rejection setups, the creation of the training data is different for the two. We describe the training strategy for the two schemes below.

#### 4.1.1. Training domain decoupled R1 rejection modules

In order to train a domain specific rejection modules, we first decode the training set through the corresponding domain's SLU models. This yields domain specific hypothesis and we compare it to the SLU annotation (which can belong to other domain). The training label for the domain specific hypothesis is set to "Accept" in case the hypothesis matches the annotation, or "Reject" otherwise. Note that in this case, the word embeddings are learnt over the vocabulary of words in the training data, while the vocabulary of hypothesis (domain, intent and slots) is limited to a single domain's labels. The training data for a given rejection module in this scheme will have a high label imbalance as hypotheses for utterances belonging to other domains will always carry a "Reject" label.

#### 4.1.2. Training system level R2 rejection module

The training set for this rejection module is generated post SLU decoding from all the domains. The top SLU hypothesis post pooling is considered for acceptance/rejection. The training label for the top hypothesis is set to "Accept" in case the hypothesis matches the annotation, or "Reject" otherwise. While the vocabulary of words in the training data is same as the vocabularies for each R1 rejection module, the vocabulary of hypothesis in this case consists of all domains, intents and slots an SLU system can produce. We expect to see more "Accept" labels in the training data as compared to the modules in the R1 scheme. We next define the FAR and FRR metrics used to evaluate the rejection modules.

### 4.2. FAR and FRR definition

In case of the scheme R2, we define a false accept and false reject in the conventional sense (an accepted incorrect top hypothesis is a false accept, while a rejected correct top hypothesis is a false reject). The count of false accepts/rejects are divided by the test set size to get FAR/FRR.

In case of the scheme R1, we compute FAR and FRR at two stages in the SLU system: once for each of the rejection modules and then for the overall system. The FAR and FRR values are again computed in the conventional sense for the top domain specific hypothesis per R1 module. Post the pooling of accepted hypothesis, we compute another set of FAR/FRR values. The count of false accepts is the number of test instances with an incorrectly accepted top hypothesis. The count of false rejects is the number of instances where the correct hypothesis is a candidate hypothesis, but is rejected along with all other candidate hypotheses. This definition of false rejects is in parity with the definition in the R2 scheme, as rejection of a correct top hypothesis in R2 also leads to SLU system not returning any hypothesis. For each FAR/FRR result, we also compute an overall F1-error metric by computing their harmonic mean.

## 5. EXPERIMENTAL SETUP

We conduct two sets of experiments given the modeling architecture in the previous section. In the first experiment, we compare the performances of the decoupled and system level rejection modules on an SLU system trained on oracle human generated transcripts. We present our analysis and consideration in choosing one architecture over the other. In the second experiment, we evaluate the performance of the rejection modules in a real world scenario where the SLU system operates on transcripts generated by an ASR system.

### 5.1. Hypotheses Rejection on human generated transcripts

In this experiment, we obtain an SLU system trained on human transcribed utterances. We use a set of ∼10M utterances directed to a smart speaker device to train an SLU system as described in Section 2. The DC, IC and NER components are trained on one section of the data, followed by re-ranker training on a separate held out section. We use another set of ∼720k utterances to train the rejection modules and ∼100k utterances as development set. Each of these utterances are fed through the entire SLU system and the generated hypotheses are used to train all the rejection modules in R1 and R2. Finally, a separate set of ∼100k utterances are used as a test set.

**Table 1**. FAR, FRR and F1 metrics using various feature sources in the R1 and R2 rejection schemes.

| **Features used** | **FAR** | **FRR** | **F1** |
|---|---|---|---|
| **No rejection module** | | | |
| - | 10.9 | 0.0 | - |
| **R1: Domain specific rejection** | | | |
| SLU score | 4.9 | 5.0 | 5.0 |
| + Utterance embedding | 5.5 | 2.5 | 3.4 |
| **+ SLU hypothesis embedding** | 4.5 | 2.5 | **3.2** |
| **R2: System level rejection** | | | |
| SLU score | 7.1 | 3.1 | 4.3 |
| + Utterance embedding | 5.2 | 3.2 | 3.9 |
| **+ SLU hypothesis embedding** | 4.6 | 2.5 | **3.2** |

#### 5.1.1. Results

Table 1 provides the results on the FAR and FRR metrics for the two rejection schemes. The first result in the table is considering an SLU system which does not contain any rejection module. Once a top hypothesis is generated, it is necessarily sent to downstream components. In that case, we observe that about 10.9% of the utterances in the test set carry an incorrect top hypothesis – setting an equivalent FAR. The FRR is listed as 0% as none of the correct top hypothesis are rejected. Using just the SLU confidence for classification is akin to finding a Bayes optimal boundary based on a single feature [16]. This is akin to the typical setting in conventional machine learning, where rejection is done using only the confidence score. However, with addition of more features, we obtain better performance in terms of F1 error metric. The improvement in rejection with availability of more features is intuitive, as we train the rejection model on a separate dataset that helps it infer if/when to accept a "low" SLU score hypothesis conditional on the utterance and the associated hypothesis (and vice-versa). We also do not observe any significant difference between the best performing R1 and R2 schemes, showcasing that either of the schemes could be chosen as far as the model performance is concerned.

#### 5.1.2. Discussion

Our experiments suggest that the schemes R1 and R2 carry similar power in terms of overall rejection performance. However, the schemes are inherently different in certain respects. We note that the following differences between the designs of R1 and R2:

•The scheme R1 consists of as many rejection modules as the number of domains, while R2 only requires training one rejection module. While scheme R1 may add to operational burden of maintaining multiple rejection modules, it also allows for a divergence in architecture, features as well as datasets used in training the rejection modules. If a distributed model ownership and maintenance is a consideration (e.g. in an organization with several teams), scheme R1 is more applicable.

•In case of R2, a change in any domain's models will require retraining the rejection module as change in any domain's models will impact the overall top SLU hypothesis. On the other hand, change to a given domain's artifacts only requires retraining that domain's rejection module in R1.

•Once a correct hypothesis is falsely rejected in either schemes R1 or R2, it can not be recovered. On the other hand, a hypothesis falsely accepted by a module in R1 may not be an overall false accept - as it may be superseded by another domain's hypothesis post the pooling of accepted hypotheses. For the same reason, a true accept in R1 by a module is not guaranteed to be overall true accept. In case of R2, once a top hypothesis is falsely/correctly accepted, it is guaranteed to be a false/true accept.

**Table 2**. Examples of errors committed by the rejection module in the absence of ASR features. Words in bold show the ASR errors. Only the example in Case C is corrected after the addition of ASR features.

| Case | ASR Text | SLU hypothesis correct? | Rejection module | |
|---|---|---|---|---|
| | | | with ASR features | without ASR features |
| A | Check my schedule | Yes | Reject | Reject |
| B | What **he** (insertion) show do you recommend | Yes | Reject | Reject |
| C | Turn my ligts **on** (substitution - to) orange | No | Accept | Reject |
| D | Answer message | No | Accept | Accept |

**Table 3**. FAR, FRR and F1 metrics on a R1 rejection module operating on an ASR + SLU system.

| **Features used** | **FAR** | **FRR** | **F1** |
|---|---|---|---|
| **No rejection module** | | | |
| - | 9.9 | 0.0 | - |
| **R1: Domain specific rejection** | | | |
| SLU score | 4.5 | 4.8 | 4.6 |
| + Utterance embedding | 5.1 | 2.7 | 3.5 |
| + SLU hypothesis embedding | 4.2 | 2.3 | 3.0 |
| + **ASR features** | 3.8 | 1.9 | **2.6** |

•Another difference is that in R1, each domain can tune it's operating point for FAR/FRR trade-off independent of other domains.

### 5.2. Hypotheses Rejection on ASR generated transcripts

SLU systems are often trained on human generated transcripts, but integrated with an ASR in real time to operate on machine generated text. This practice is typically followed for the following two reasons: (i) it allows for a decoupled training of ASR and SLU models and, (ii) annotating ASR generated transcripts for SLU model training is difficult as it is not clear as to how to annotate ASR errors with SLU training labels. However, after training an SLU system on human generated transcripts, a rejection module can be trained on an SLU system operating with ASR hypothesis. Such a rejection module can also learn to reject incorrect hypothesis arising due to ASR errors. The objective of this experiment is to train a rejection module on features extracted from SLU and ASR systems and analyze the decisions of the rejection module in presence of ASR errors. We borrow the same architecture as described in Section 4 and add ASR features along with the ASR generated text (1-best), SLU score and SLU hypothesis. Same test data as described in Section 5.1 is used for evaluation, however the text for the data is generated by an ASR system [17] (due to missing audio for a portion of the test set, the test set in this experiment is about 60% of the original test set).

#### *5.2.1. ASR features*

We use the following features to train the rejection module (i) product of confidence scores of all tokens (including eps) in the 1-best path of the confusion network, (ii) product of posteriors of all tokens (including eps) in the 1-best path of the confusion network (iii) logarithm of the number of frames in the utterance, (iv) logarithm of the number of words in 1-best confusion network hypothesis and, (v) logarithm of the mean of the number of arcs explored at each frame.

#### *5.2.2. Results*

The results for the rejection module applied on SLU system operation on ASR transcripts is shown in Table 3. We only show results for the scheme R1 as results were similar with the scheme R2. Note that the FAR/FRR values are different than those reported in Table 1 as only a subset of test data audio was available for ASR decoding and now the rejection module is operating on ASR generated text. Expectedly, factoring in ASR features yields better FAR/FRR values. In the next section, we perform analysis on the rejected hypothesis.

#### *5.2.3. Discussion*

We note that irrespective of the correctness of the ASR text, the rejection module should make an "accept" decision as long as the SLU hypothesis is correct. Table 2 shows (synthetic) examples of decisions made by the rejection module before and after the addition of ASR score feature. We observe that the introduction of ASR feature leads to better rejection of erroneous hypothesis arising due to ASR errors (case C). However, we did not observe an increased acceptance of the instances falling in the case B category. We expect an edited ASR feature that pays attention to words important for SLU (for instance the word 'he' is not important for SLU in case B) can be useful to alleviate such errors. Overall, training a rejection module on SLU and ASR features still allows for a decoupled training of SLU and ASR modules while learning to reject errors that arise due to an erroneous interaction between the two components. Such a scheme requires retraining the rejection modules post the SLU/ASR update, which is cheaper in practice when compared to a coupled training of SLU and ASR models.

## 6. CONCLUSION

Given that *no response* is sometimes better than *incorrect response*, hypothesis rejection modules in an SLU system can enhance experience. In this work, we explore two schemes for rejecting incorrect SLU hypotheses: rejecting hypotheses produced by domain specific components (scheme R1) and rejecting hypotheses at the overall SLU system level (scheme R2). We experiment with input features such as the utterance and SLU hypothesis, in addition to the SLU confidence score, and observe that more information can improve the rejection performance. We observe that while the schemes R1 and R2 provide similar performance capabilities, they carry some inherent differences which should be considered while choosing one over the other. We also experiment with rejection modules trained on SLU and ASR based features and observe that the rejection module is capable of identifying SLU errors arising due to a faulty ASR transcription.

In the future, we aim to experiment with different modeling architectures and by adding more features to the rejection modules. One particular experimental setup could involve assessing the overall FAR/FRR impact of re-training the R1 rejection modules independently. This experiment will involve updating a domain specific R1 rejection module independent of others (e.g. using different data, model configurations, changes in domain annotation schema etc.). Finally, we aim to experiment with end-to-end SLU setup [18] and extend the concept of hypothesis rejection. One experimental setting in end-to-end SLU systems will involve adding a separate loss component for rejection to the overall optimization.